\documentclass{article}

\usepackage{microtype}
\usepackage{graphicx}
\usepackage{subfigure}
\usepackage{booktabs} 

\usepackage{color, colortbl}
\definecolor{lightgray}{rgb}{0.78125,0.78125,0.78125}
\definecolor{darkgray}{rgb}{0.46875,0.46875,0.46875}
\usepackage{siunitx} 
\DeclareSIUnit\px{px}
\DeclareSIUnit\cell{cell}

\usepackage{hyperref} 

\usepackage[accepted]{icml2020}

\icmltitlerunning{C2G-Net: Exploiting Morphological Properties for Image Classification}

\begin{document}

\twocolumn[
\icmltitle{C2G-Net: Exploiting Morphological Properties for Image Classification}


\icmlsetsymbol{equal}{*}

\begin{icmlauthorlist}
\icmlauthor{Laurin Herbsthofer}{cbmed} 
\icmlauthor{Barbara Prietl}{cbmed,endo}
\icmlauthor{Martina Tomberger}{cbmed}
\icmlauthor{Thomas Pieber}{cbmed,endo,joanneum}
\icmlauthor{Pablo L\'{o}pez-Garc\'{i}a}{cbmed,imi}
\end{icmlauthorlist}

\icmlaffiliation{cbmed}{Center for Biomarker Research in Medicine GmbH, Graz, Austria}
\icmlaffiliation{imi}{Institute for Medical Informatics, Statistics and Documentation, Medical University of Graz, Graz, Austria}
\icmlaffiliation{endo}{Division of Endocrinology and Diabetology, Medical University of Graz, Graz, Austria}
\icmlaffiliation{joanneum}{Joanneum Research Forschungsgesellschaft mbH, Health Institute for Biomedicine and Health Sciences, Graz, Austria}

\icmlcorrespondingauthor{Laurin Herbsthofer}{laurin.herbsthofer@cbmed.at}

\icmlkeywords{Machine Learning, ICML, image, compression, micrograph, segmentation, object detection, morphology, CNN, immunohistochemistry, IHC, biology, cell, FFPE, staining}

\vskip 0.3in
]



\printAffiliationsAndNotice{}  

\begin{abstract}
In this paper we propose C2G-Net, a pipeline for image classification that exploits the morphological properties of images containing a large number of similar objects like biological cells.
C2G-Net consists of two components: (1) Cell2Grid, an image compression algorithm that identifies objects using segmentation and arranges them on a grid, and (2) DeepLNiNo, a CNN architecture with less than 10,000 trainable parameters aimed at facilitating model interpretability.
To test the performance of C2G-Net we used multiplex immunohistochemistry images for predicting relapse risk in colon cancer.
Compared to conventional CNN architectures trained on raw images, C2G-Net achieved similar prediction accuracy while training time was reduced by 85\% and its model was is easier to interpret.
\end{abstract}

\section{Introduction}
\label{intro}
Convolutional neural networks (CNNs) have revolutionized image analysis and are the de facto standard for image classification tasks \cite{Krizhevsky2012-us,Russakovsky2015-yl}. However, CNNs still suffer from practical limitations in many applications as the size of training images increases and network architectures become deeper \cite{He2015-uo,Russakovsky2015-yl}. Due to limitations in computational resources, large images used for training usually need to be downscaled or tiled, larger strides of the convolutional and pooling layers employed, smaller batch sizes or lower precision calculations used, etc \cite{Wang2015-dh,Goodfellow2016-cw,Liu2017-fc,Mungle2017-pd,Komura2018-xv,Pinckaers2018-lz,Ozge_Unel2019-vr}. In addition, models typically require long training times and may be difficult to interpret for humans. Reasonable training times and interpretable models, however, are important requirements in applications such as health care \cite{Ratner2018-js}.

Especially when dealing with biological tissue, images often exhibit some special morphological properties. As an example, a micrograph of a biological tissue section at 40x magnification will show the individual biological cells that constitute the tissue \cite{Stack2014-uc,Parra2019-yy,Hofman2019-uu}. At a resolution of \SI{0.5}{\micro\metre}, a single biological cell with a diameter of \SI{10}{\micro\metre} is captured by several hundred pixels in the image. For many biological and clinical questions, this sub-cell information is less relevant than the phenotype of the cells and their relative location in the tissue. Furthermore, a whole-slide-image (WSI) of a single tissue slide with a diameter of \SI{1}{\centi\metre} can be several Gigabytes large, while only including about a million biological cells.

CNNs have been used extensively with these images, including hematoxylin-eosin (HE)-stained tissue slides \cite{Kather2019-pa}, as well as (multiplex) immunohistochemistry (IHC) stainings \cite{Xu2016-sn,Khosravi2018-qc,Gertych2019-oy,Bulten2019-qt}. However, to the best of our knowledge, it has not been thoroughly studied how the morphological properties of these images can be exploited in CNNs to reduce training time and increase model interpretability.  

In this paper we propose C2G-Net, a pipeline for image classification that consists of (1) Cell2Grid, an image compression algorithm that exploits morphological information in images through segmentation, and (2) DeepLNiNo, a compact CNN architecture that provides an interpretable model.

\section{C2G-Net pipeline}
\label{method}
\begin{figure*}[]
\begin{center}
\centerline{\includegraphics[width=\textwidth]{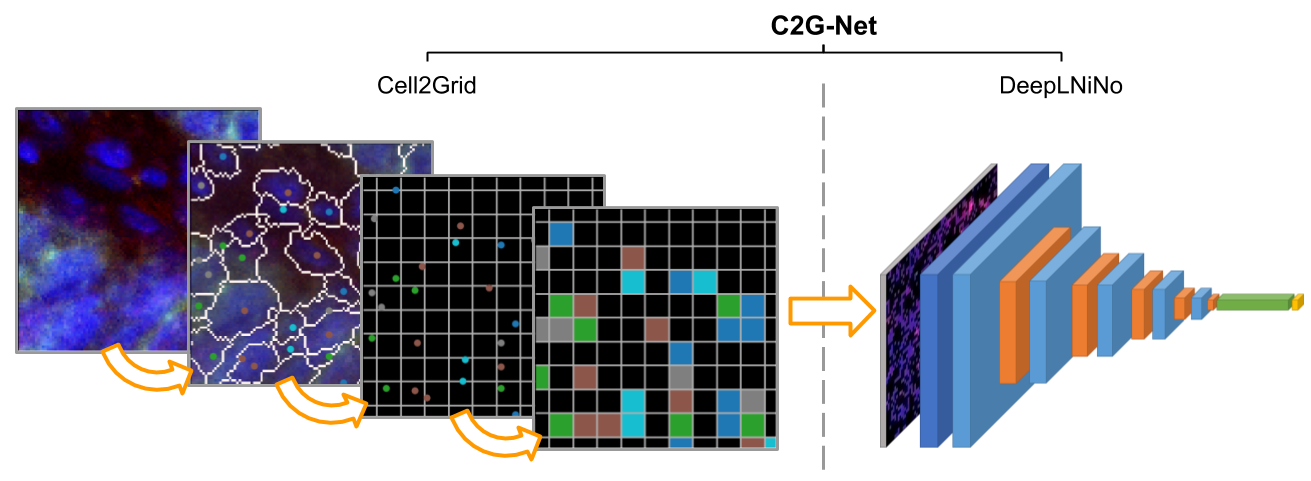}}
\caption{C2G-Net consisting of: (1) Cell2Grid, an image compression algorithm that significantly reduces training time and (2) DeepLNiNo, a CNN architecture designed to operate on Cell2Grid-compressed images and improve interpretability.}
\label{C2G-Net-Abstract}
\end{center}
\end{figure*}

In this section we describe our method, C2G-Net (see \autoref{C2G-Net-Abstract}), which consists of two components: Cell2Grid, an image compression algorithm that exploits the morphological properties of a specific class of images and DeepLNiNo, a CNN architecture designed to operate on these compressed images. Our method is applicable to images that contain a large number of similar objects.

\subsection{Cell2Grid image compression algorithm}
The Cell2Grid image compression algorithm consists of the three steps listed below.

\textbf{Step 1: Object identification} \\
During the first step, relevant objects (e.g. biological cells) are identified in the image. For each object, its location within the image as well as other relevant properties for the image analysis task (e.g. average color, size, or shape of the object) are extracted. The final output of the object identification step is therefore a list of objects with their associated xy-coordinates $\vec{X}$ and properties, similar to point-cloud data.

\textbf{Step 2: Assigning objects to target grid} \\
The goals of Step 2 are to first obtain a target grid that is virtually placed over the images and then place objects from Step 1 on that target grid.

The target grid is a square grid with grid spacing $d$ in units of image pixels and it should be as coarse as possible (to achieve a high compression ratio) while simultaneously being as fine as necessary to prevent that grid locations become overpopulated by objects (i.e., an excessive amount of assignment conflicts).

As a simple heuristic for estimating a suitable target grid spacing $d$, we use
\begin{equation} \label{eq:1}
d \approx \frac{1}{2} \left(\frac{1}{n} \sum_{i=1}^{n} \sqrt{1/\rho_i}\right),
\end{equation}
where $\rho_i$ is the object density of image $i$ and the sum includes all images to be compressed in a single batch. The square root in \autoref{eq:1} transforms the inverse of the object density (the average area occupied by each object) into a spatial dimension, estimating the average object extent. After averaging over all $n$ images in the batch, we divide by $2$ to account for possible empty space in the images and local density variations. If desired, the computed value can be rounded to the next integer to ease explainability and visualization of the compressed images.

Once grid spacing is calculated, all identified objects are assigned to the target grid by binning their original xy-coordinates $\vec{X}$ to the discrete nodes of the target grid: 
\begin{equation} \label{eq:2}
\vec{X}_{g} = \lfloor \vec{X}/d \rceil,
\end{equation}
whereas the $\lfloor\cdot\rceil$ denotes conventional rounding to the next integer and $\vec{X}_{g}$ corresponds to the coordinates of pixels in the final compressed image.

If two objects are within a distance of $d\sqrt{2}$ they may be assigned to the same $\vec{X}_{g}$. To achieve a one-to-one relation of objects to pixels, we require that each object is assigned uniquely to a grid node. Therefore, possible assignment conflicts are resolved by either moving objects to free adjacent grid nodes or by deleting them from the data. In \autoref{AppendixA}, we introduce \textit{PriorityShift}, a simple algorithm that resolves conflicts locally in a computationally efficient way. After this step, nodes in the target grid are either empty or contain exactly one object and its associated properties.

\textbf{Step 3: Image compression} \\
In Step 3, a low-resolution compressed output image is produced. After Step 2, grid nodes contain the properties of assigned objects, similarly to how pixels contain RGB color values in conventional images. In our case, the resulting data structure is a tensor of size $K_x\times K_y\times P$, where $K_x$ and $K_y$ are the number of grid lines in x and y dimension, respectively, and $P$ is the number of object features stored at each node. Grid nodes without objects are assigned zero as a default value for every object property. Since a square grid was used, our tensor can be converted into an image with $P$ color channels, e.g. by using a lossless image format with multiple channels, like multi-channel TIFF. This multiplex image, which we term Cell2Grid image, is the final output of the image compression step.

An example of the whole process for a biological image from our case study is introduced in \autoref{sec:materials} (\autoref{fig:c2g_process}).

\subsection{DeepLNiNo CNN architecture}
To exploit the special properties of Cell2Grid images (further information in the Discussion section) we experimented with different CNN architectures that are designed for spatially sparse input, with  DeepCNet \cite{Graham2014-zm} and the Network-in-Network approach \cite{Lin2013-wk} being the most relevant for our task and the ones we built on. A \verb!DeepCNet(l,k)! network consists of $l$ convolutional layers, each of which is followed by a max pooling layer. The number of feature maps in the convolutional layers increases from layer to layer, being $n\times k$ in layer $n$. While DeepCNet consists of one layer with filter size $3\times3$ followed by several layers with filter size $2\times2$, the Network-in-Network architecture also makes use of convolutional layers with filter size $1\times1$.

\begin{table}[h]
\caption{DeepLNiNo architecture.}
\label{table:DeepLNiNo} 
\vskip 0.1in
\begin{center}
\begin{small}
\begin{tabular}{lrrr}
\toprule
\begin{sc}type\end{sc} & \begin{sc}filter size\end{sc} & \begin{sc}output size\end{sc} & \begin{sc}params\end{sc} \\ 
\midrule
input 			& 				& $135\times101\times6$ 	& 		\\
conv 			& $1\times1$ 	& $135\times101\times16$ 	& 112 	\\
conv 			& $2\times2$ 	& $134\times100\times16$ 	& 1,040 	\\
maxpool 		& $2\times2$ 	& $67\times50\times16$ 		& 		\\
conv 			& $3\times3$ 	& $65\times48\times16$ 		& 2,320	\\
maxpool 		& $3\times3$ 	& $21\times16\times16$ 		& 		\\
conv 			& $3\times3$ 	& $19\times14\times16$ 		& 2,320	\\
maxpool 		& $3\times3$ 	& $6\times4\times16$ 		& 		\\
conv 			& $3\times3$ 	& $6\times4\times16$ 		& 2,320	\\
maxpool 		& $3\times3$ 	& $2\times2\times16$ 		& 		\\
conv 			& $2\times2$ 	& $2\times2\times16$ 		& 1,040	\\
maxpool 		& $2\times2$ 	& $1\times1\times16$ 		& 		\\
flatten 		& 				&  							& 		\\
dense 			& 				& $1\times1\times32$ 		& 544	\\
dropout 		& 33\%			&							& 		\\
softmax 		& 				& $1\times1\times2$ 		& 66	\\
\midrule
params total	& 				& 							& \textbf{9,762} \\
\bottomrule
\end{tabular}
\end{small}
\end{center}
\end{table}

Our goal was to create a CNN architecture that has significantly fewer trainable parameters than comparable architectures and is easy to interpret while still performing well in a binary image classification task. For that purpose, we introduce Deep L1-regularized Network-in-Network narrOw (DeepLNiNo) architecture, an easy-to-interpret CNN designed to work with Cell2Grid images of size \SI{135x101}{\px} and six color channels. DeepLNiNo is small (less than 10,000 trainable parameters), sparse (L1-regularization of the first convolutional layer with $1\times1$-kernel) and narrow (only 16 filters per convolutional layer). \autoref{table:DeepLNiNo} lists the network architecture.

In DeepLNiNo, each convolutional layer has 16 filters, with the first one using a kernel size of $1\times1$, as inspired by the Network-in-Network approach. This layer is L1-regularized to introduce sparsity and increase model interpretability. Since one pixel in Cell2Grid images corresponds exactly to one object, this layer enables the tracking of learned object properties. The remaining convolutional layers use a mix of kernel sizes and image padding methods. Dropout \cite{Hinton2012-ft} is applied only after the fully connected dense layer. In total, DeepLNiNo consists of 9,762 trainable parameters.

\subsection{Reference model}
As a reference CNN architecture, we used a modification of the DeepCNet architecture proposed by Graham \yrcite{Graham2014-zm} that was extended by a fully connected layer with 128 neurons at the end of the network. We used the parameters $l=6$ and $k=32$, resulting in a network with six convolutional layers, and $32\times n$ feature maps in convolutional layer $n$. We denote this model as \verb!DeepCNet(l=6)! indicating the number of layers used while keeping $k=32$ for all models.

\section{Case Study: Colon Cancer Relapse}
\label{experiment}
To test the performance of C2G-Net, we used a real use case from health care: predicting the relapse risk of colon cancer patients. In a retrospective study, surgically removed tumor tissue samples from 48 colon cancer patients were collected from a biobank. Each patient was observed for at least 3 years after tumor surgery and their final tumor recurrence status labeled as either \textit{relapse} (12 patients, 25\%) or \textit{no-relapse} (36 patients, 75\%). The goal was to predict the final tumor recurrence status (i.e. relapse/no-relapse) for each single image.

We wanted to investigate (a) how two different CNN model architectures perform on Cell2Grid images, (b) how the three different types of input images (raw, RGB and Cell2Grid) at the same image resolution influence the performance of a single CNN model and (c) how training time differs for the best performing models on all three input image types.

\subsection{Materials}
\label{sec:materials}
\begin{figure*}[]
\vskip 0.1in
\begin{center}
\centerline{\includegraphics[width=0.78\textwidth]{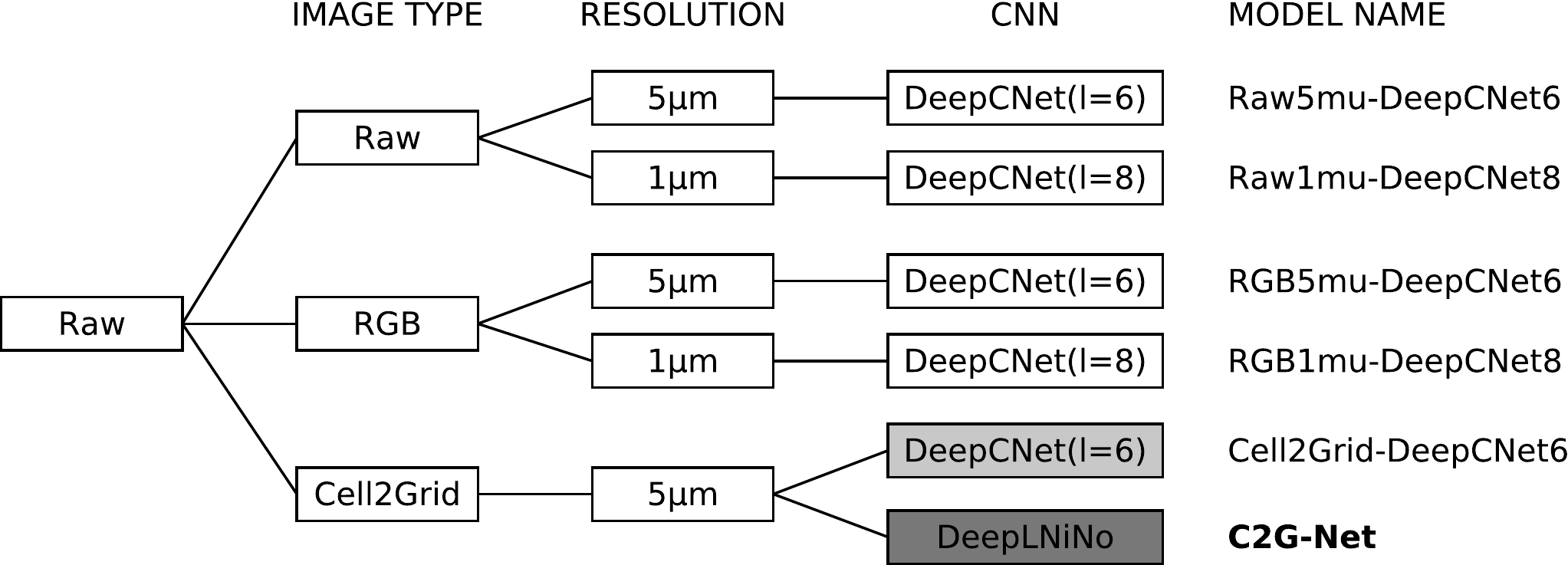}}
\caption{The six different model configurations used in our experiment.}
\label{fig:eval_setup}
\end{center}
\vskip -0.25in
\end{figure*}

Images consisted of multiplex immunohistochemistry (mIHC) micrographs of formalin-fixed paraffin embedded (FFPE) tissue sections from the tumor \cite{Fox1985-jf} stained with six fluorescence-conjugated antibodies (CD3, CD8, CD45RO, PD-L1, FoxP3, Her2 plus DAPI for staining the nuclei of the cells) \cite{Ramos-Vara2014-fw}. Images covered an area of \SI{672x504}{\micro\metre} and were recorded at 40x magnification, with a final size of \SI{1344x1008}{\px} at a \SI{0.5}{\micro\metre/\px} resolution and six color channels, one for each antibody used. 1,119 mIHC images from the 48 patients were recorded in total. These images are referred to as \textit{raw} images.

\begin{figure*}[p]
     \centering
     \begin{subfigure}[]{}
         \centering
         \includegraphics[width=.23\textwidth]{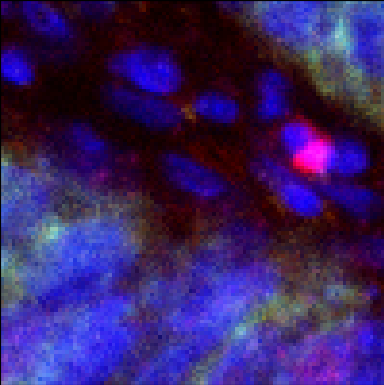}
         \label{fig:c2g_a}
     \end{subfigure}
     \hfill
     \begin{subfigure}[]{}
         \centering
         \includegraphics[width=.23\textwidth]{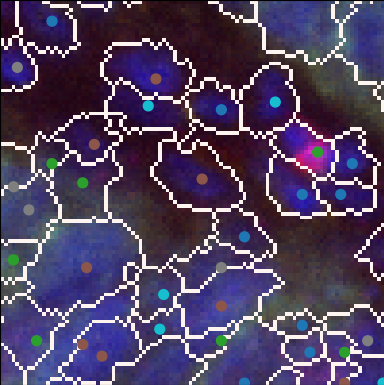}
         \label{fig:c2g_b}
     \end{subfigure}
     \hfill
     \begin{subfigure}[]{}
         \centering
         \includegraphics[width=.23\textwidth]{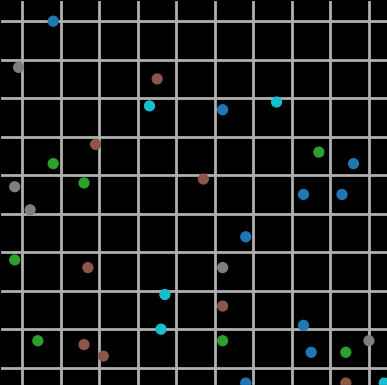}
         \label{fig:c2g_c}
     \end{subfigure}
     \hfill
     \begin{subfigure}[]{}
         \centering
         \includegraphics[width=.23\textwidth]{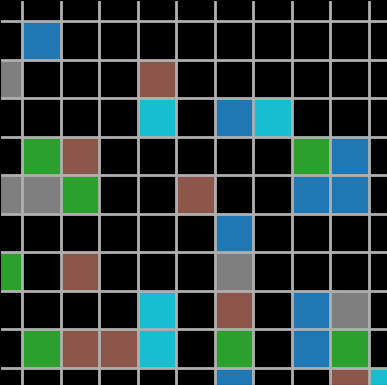}
         \label{fig:c2g_d}
     \end{subfigure}
        \caption{Cell2Grid image compression procedure and final outcome. (a) Small part of a raw image, visualized as RGB image ($100\times100$ pixels covering an area of \SI{50x50}{\micro\metre}); (b) outlines of the identified objects (biological cells) in white and dots with random colors showing the locations of the cells nuclei; (c) point-like locations of objects with superimposed target grid with grid spacing $d=$ \SI{5}{\micro\metre}; (d) final Cell2Grid image ($10\times10$ pixels); assignment conflicts (e.g. green and gray cell in the bottom right corner) resolved with \textit{PriorityShift}. The final Cell2Grid image is sparse, containing several completely black pixels, even in regions with many objects.}
        \label{fig:c2g_process}
\end{figure*}

Based on this set of raw images, we created two additional data sets termed {RGB} and {Cell2Grid}. RGB images are pseudo-color images where the six color channels of the raw image are mapped to distinct RGB color values. \autoref{fig:c2g_process} illustrates the compression of raw images using Cell2Grid. For object identification we used inForm \cite{Akoya_Biosciences_undated-lg}, a commercially available software for biological cell segmentation. On average, images contained 4,149 biological cells in an area of \SI{338688}{\micro\metre\squared} ($\approx$\SI{81.6}{\micro\metre\squared/\cell}). For all six recorded IHC markers, the average color channel intensities (i.e. marker expressions) over the entire cell area were calculated and used as object properties. We calculated the target grid spacing according to \autoref{eq:1} using all images to be $d=$ \SI{4.7}{\micro\metre} $\approx$ \SI{5}{\micro\metre}. This is comparable to the size of a lymphocyte \cite{Wood2004-vv}. We used the location of the cells nuclei as point-like object locations for assignment to the grid. Assignment conflicts were resolved with \textit{PriorityShift}.

\subsection{Methods and Evaluation}

We created 6 different models by combining image type (raw, RGB, and Cell2Grid), resolution (\SI{1}{\micro\metre} and \SI{5}{\micro\metre}), and CNN (DeepCNet and DeepLNiNo)---see \autoref{fig:eval_setup}. Raw and RGB images were rescaled to \SI{5}{\micro\metre} and \SI{1}{\micro\metre} using bilinear interpolation \cite{Clark2015-go} and the number of layers in DeepCNet  adapted to accommodate for image size. We implemented all models in Python using Keras \cite{Chollet2015-nn} and trained on a conventional desktop workstation with one GPU (Nvidia RTX 2070).

Each model was trained 10 times to estimate model variance and all $n=1,119$ images split into a training and a validation set ($2/3$ and $1/3$ of images, respectively) by stratifying samples based on their relapse class. Training images were oversampled to account for class imbalance. To further emphasize the clinical importance of predicting the relapse class, we weighted corresponding samples by 3:1 in favour of relapse, and used standard data augmentation to avoid overfitting. For Cell2Grid images, we had to implement our own data augmentation methods due to the special properties of Cell2Grid images (see \autoref{AppendixB} for details). Cross entropy was used as loss function and Adadelta \cite{Zeiler2012-ov} as optimizer. 

\subsection{Results}
\begin{figure*}[p]
\begin{center}
\centerline{\includegraphics[width=0.9\textwidth]{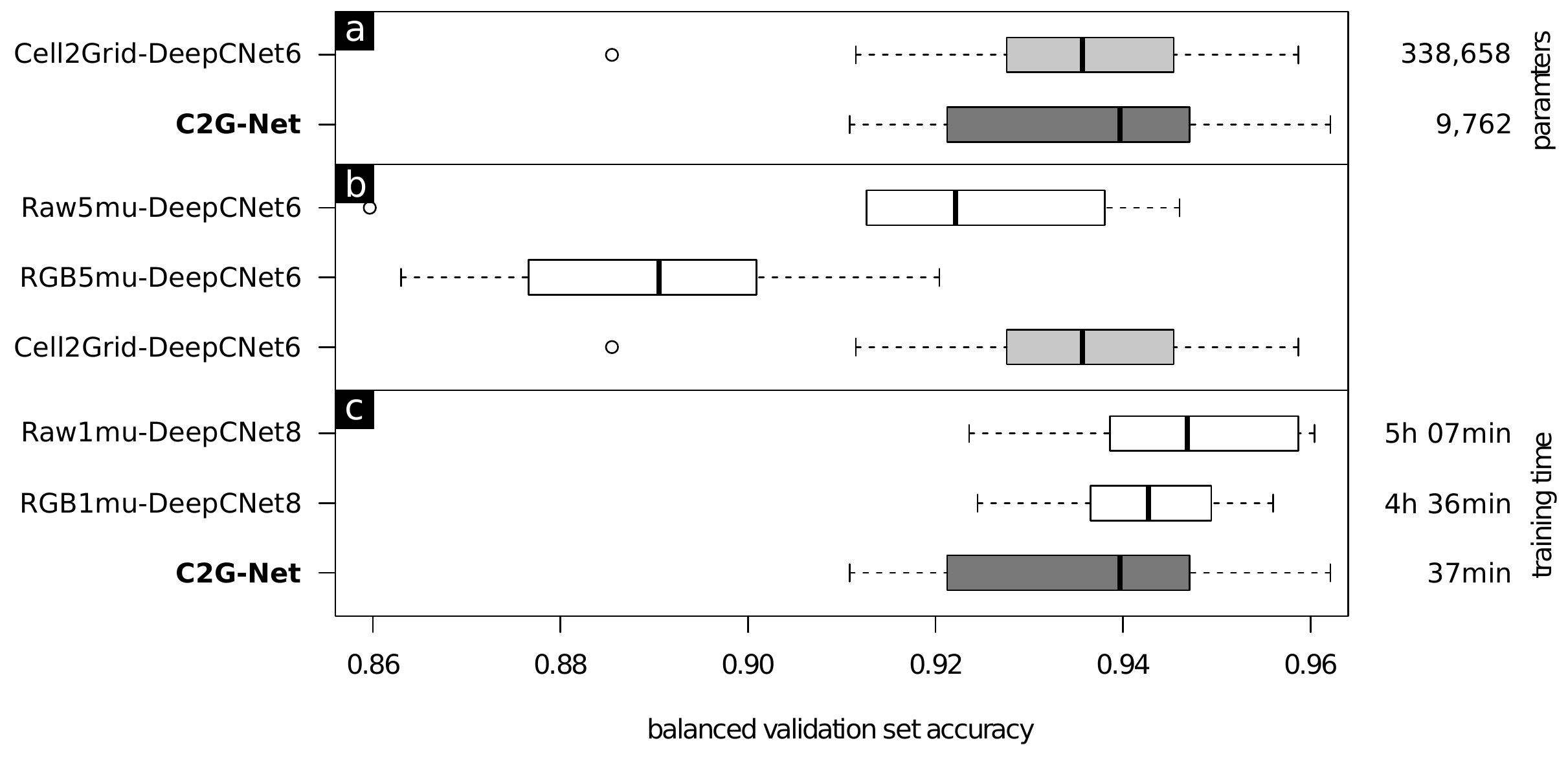}}
\caption{Balanced validation set accuracy of 10 model runs each, see \autoref{results-table}. (a) Comparison of CNN architectures using Cell2Grid images; (b) Comparison of different input data types using DeepCNet architecture; (c) comparison of best models for each input data type. C2G-Net in dark gray, Cell2Grid-DeepCNet in light gray.}
\label{valsetaccuracy}
\end{center}
\end{figure*}

We experimented with the number of training epochs and found that all DeepCNet models fully converged after 400 epoch while the DeepLNiNo model required 1000 epochs, likely an effect of L1-regularization. \autoref{results-table} shows the mean and standard deviation for balanced validation set accuracy and training times for the 10 runs per model, results are illustrated in \autoref{valsetaccuracy}. \autoref{valsetaccuracy}a shows that both DeepLNiNo and DeepCNet perform equally well when trained on Cell2Grid images, although DeepLNiNo contains significantly fewer trainable parameters.
Using the same DeepCNet architecture with different input data types, \autoref{valsetaccuracy}b shows that using Cell2Grid images leads to the most accurate models.

Finally, \autoref{valsetaccuracy}c compares the best models for each input data type. While C2G-Net can not outperform the models using data at \SI{1}{\micro\metre} resolution, the required training time was reduced by more than a 85\%.
As expected, a general trend towards worse performance with decreasing input image resolution can be observed for both raw and RGB images.

\begin{table*}[p]
\vskip 0.3in
\caption{Mean (standard deviation) for balanced validation set accuracy for 10 model runs and average training time of a single run for different CNN architectures and input data types. Image resolution denotes the resolution of images for training after compression.}
\vskip 0.1in
\label{results-table}
\begin{center}
\begin{small}
\begin{tabular}{llrlrr}
\toprule
\begin{sc}Model name\end{sc} & \begin{sc}Image type\end{sc} & \begin{sc}Image res.\end{sc} & \begin{sc}CNN\end{sc} & \begin{sc}Bal. accuracy\end{sc} & \begin{sc}Training time\end{sc}  \\ 
\midrule
Raw5mu-DeepCNet6 & Raw & \SI{5}{\micro\metre} & DeepCNet(l=6) & 0.921 (0.024) & 00:07:29 \\
Raw1mu-DeepCNet8 & Raw & \SI{1}{\micro\metre} & DeepCNet(l=8) & \textbf{0.946} (0.014) & 05:07:23 \\
RGB5mu-DeepCNet6 & RGB & \SI{5}{\micro\metre} & DeepCNet(l=6) & 0.891 (0.019) & \textbf{00:03:54} \\
RGB1mu-DeepCNet8 & RGB & \SI{1}{\micro\metre} & DeepCNet(l=8) & 0.942 (0.011) & 04:36:18 \\
\rowcolor{lightgray}
Cell2Ggrid-DeepCNet6 & Cell2Grid & \SI{5}{\micro\metre} & DeepCNet(l=6) & 0.932 (0.021) & 00:07:48 \\
\rowcolor{darkgray}
\textbf{C2G-Net} (our method) & Cell2Grid & \SI{5}{\micro\metre} & DeepLNiNo & 0.938 (0.017) & 00:37:31 \\
\bottomrule
\end{tabular}
\end{small}
\end{center}
\vskip 0.2in
\end{table*}

\section{Discussion}
\label{discussion}
In recent years, CNN architectures have become increasingly large, requiring long training times and powerful hardware, particularly when using large images for training \cite{Krizhevsky2012-us,Szegedy2014-us,Simonyan2014-bh,He2015-uo}. However, when images contain a large number of similar objects, their morphological properties can be exploited by using a compressed representation of their individual objects placed on a grid. These compressed images can then be used to train a CNN, as long as their special properties are taken into account.

\subsection{Properties of Cell2Grid images}
Cell2Grid images are different from conventional images in a number of ways: (1) they contain no natural gradients, since every pixel in the image is an independent object (e.g., a biological cell); (2) they might contain many empty pixels (numerically zero) in areas without assigned objects during compression; and (3) they can not be arbitrarily rotated, zoomed, sheared or transformed for data augmentation purposes, since the value and integrity of individual pixels matter (see \autoref{AppendixB}).

As previously shown in \autoref{fig:c2g_process}, the final Cell2Grid image is sparse, containing several completely black pixels, even in regions with many objects.
While cells surrounded by background tend to vanish in conventional downscaling due to averaging with background pixels, they are conserved during Cell2Grid compression. In this example, the Cell2Grid image has 10-times fewer pixels in each spatial direction, corresponding to a compression ratio of 100.

The Cell2Grid image compression algorithm is related to the concept of superpixels \cite{Bechar2019-uf,Akbar2015-rr}. However, while superpixels are typically used to segment the whole image into regions of similarity (including background and irrelevant objects), Cell2Grid focuses on relevant objects only and keeps their integrity intact. Compared to Voronoi tessellation \cite{Chen2017-vn} and graph-based methods \cite{Chen2017-vn} based on superpixels, our proposed places all identified objects on a target grid by approximating the location of each object to fit the grid. The final data format is that of a conventional, multi-channel image.

\subsection{Model interpretation}
\begin{figure}[!t]
\vskip 0.2in
\begin{center}
\centerline{\includegraphics[width=0.9\columnwidth]{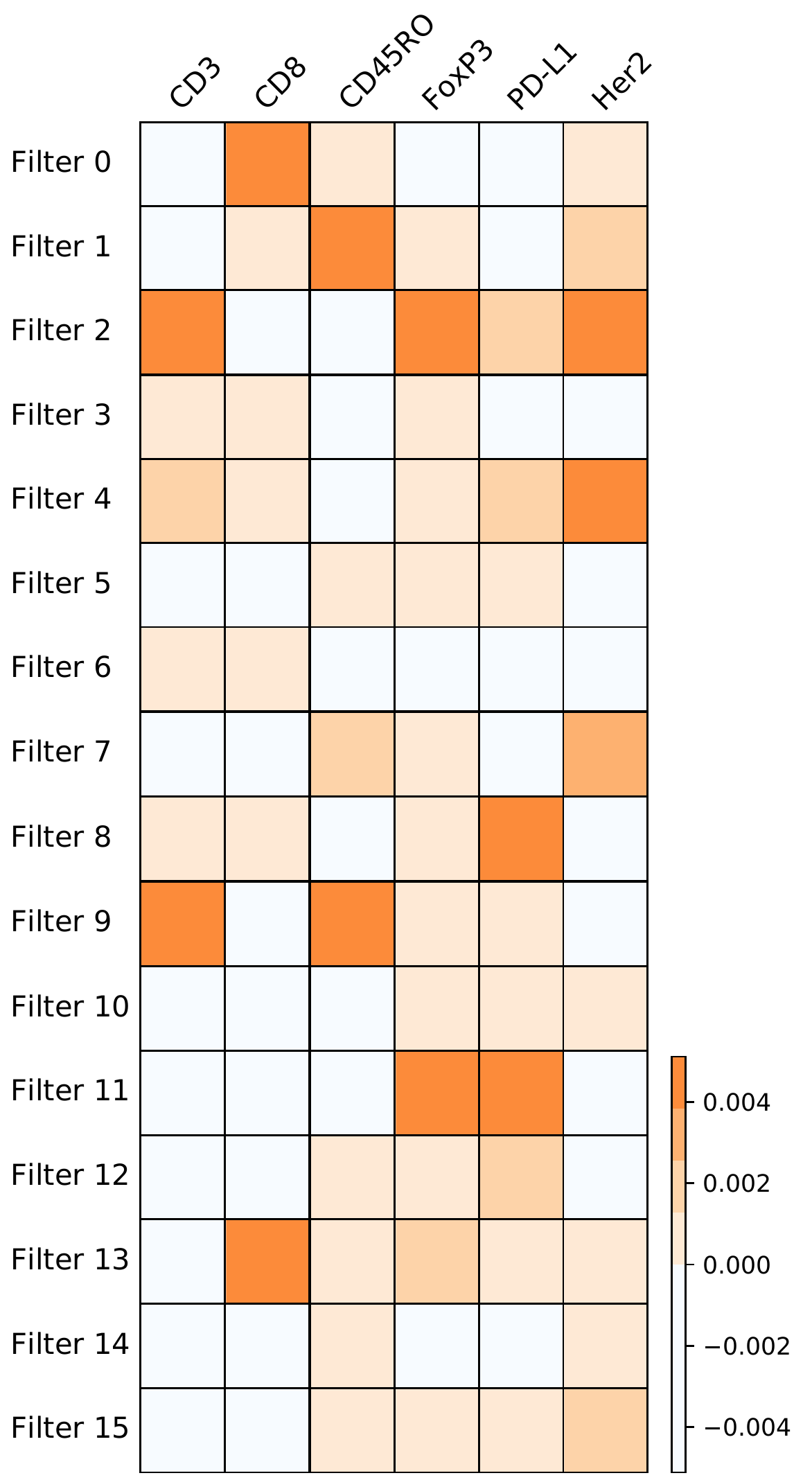}}
\caption{Learned filter weights of the first convolutional layer with kernel size $1\times1$ of a trained DeepLNiNo model. Higher values indicate increased importance of a channel for a given filter. For ease of interpretation, all negative filter weights have been colored in the same color.}
\label{fig:deeplnino_weights}
\end{center}
\vskip -0.2in
\end{figure}

Although some progress has been made towards explaining a trained CNN \cite{Zeiler2014-ni,Mahendran2015-bv,Zhang2018-hs,Ribeiro2016-bw}, keeping model architectures simple makes them easier to train and simplifies the extraction of knowledge \cite{Kaya2018-oa}. In contrast to other popular CNN model architectures that have become very deep and wide, resulting in several millions of trainable parameters \cite{Krizhevsky2012-us,Szegedy2014-us,Simonyan2014-bh,He2015-uo}, due to the narrow architecture of DeepLNiNo, the whole network and its learned weights can be displayed, facilitating interpretation.

\begin{table*}[t]
\caption{Comparison of size of training images and number of trainable parameters for C2G-Net and popular networks for image classification.}
\vskip 0.1in
\label{table:parameter_comparison}
\begin{center}
\begin{small}
\begin{tabular}{lrrr}
\toprule
\begin{sc}Model name\end{sc} & \begin{sc}Raw input image size\end{sc} & \begin{sc}File size (MB)\end{sc} & \begin{sc}Parameters\end{sc} \\
\midrule
\textbf{C2G-Net (our method)} & \SI{1344x1008x6}{} & \textbf{8.13} & \textbf{10k} \\
GoogLeNet \cite{Szegedy2014-us} & \SI{224x224x3}{} & 0.15 & 6,400k \\
ResNet-152 \cite{He2015-uo} & \SI{224x224x3}{} & 0.15 & 60,200k \\
AlexNet \cite{Krizhevsky2012-us} & \SI{224x224x3}{} & 0.15 & 62,300k \\
VGG16 \cite{Simonyan2014-bh} & \SI{224x224x3}{} & 0.15 & 138,000k \\
\bottomrule
\end{tabular}
\end{small}
\end{center}
\vskip -0.05in
\end{table*}

Since individual pixels correspond to objects in Cell2Grid images, learned filter weights in a neural network can be interpreted at an object level. \autoref{fig:deeplnino_weights} shows the learned weights of the first convolutional layer of a trained DeepLNiNo model, consisting in 16 filters with kernel size 1x1. It can be seen that Filter 9 is strongly activated by cells containing high values of CD3, CD45RO, or both. This combination of markers is characteristic for memory T-cells, which have been found previously to play a role in colon cancer prognosis \cite{Pages2005-oh}. Furthermore, every IHC marker channel appears exactly twice in all filters with a weight above $0.004$, indicating that every color channel contains useful information for classification. As expected, L1-regularization introduced some level of sparsity to the weights. Out of all filters, only eight contain at least one weight above $0.004$.

This form of model interpretation on the level of individual objects is enabled by the Cell2Grid image compression. Other image downscaling methods, like the bi-linear interpolation used for our reference data types, destroy the integrity of the biological cells by merging pixels from several cells and background together, making model interpretation more difficult. However, we acknowledge that this form of single-layer interpretation of our current DeepLNiNo architecture has its limits, since all feature maps are combined in a non-linear fashion in the downstream layers.

\autoref{table:parameter_comparison} shows the number of parameters in DeepLNiNo (10,000) compared to other popular CNNs for image classification (6-138 million). While other model architectures (e.g. VGG \cite{Simonyan2014-bh}) employ pooling much slower, DeepLNiNo applies aggressive maxpooling after every convolutional layer (except for the first). Together with the image compression step of C2G-Net, the number of parameters is kept small while still allowing for much larger input images.

\section{Future work}
In our experiment, we used raw images of size \SI{1344x1008}{\px}. However, a whole-slide-image (WSI) of a tissue section can be up to 30 times bigger in each spatial dimension. Building on our findings, we plan on applying C2G-Net to WSI data to test its performance on very large images.

Compared to conventional images, the creation of artificial Cell2Grid images is simple due to their object-like properties. In order to improve the extraction of knowledge from trained CNNs, we aim at creating simulated Cell2Grid data for which the ground truth of image labels is known by design (e.g. a difference in relative cell phenotype counts, different spatial distribution patterns etc.). We hypothesize that simulated Cell2Grid data will foster knowledge extraction from trained CNNs, which is especially important for applications in biology and health care.

When working with conventional images, local interpretable model-agnostic explanations (LIME, \cite{Ribeiro2016-bw}) have been used to break down why a CNN makes certain predictions. To this end, LIME uses superpixels \cite{Bechar2019-uf,Akbar2015-rr} and assess their contribution to the predictions. We plan on adapting LIME to work with Cell2Grid image data by occluding groups of similar objects instead of superpixels.

\section{Other applications}
In this paper we applied C2G-Net to biological data in form of mIHC images. However, we hypothesize that our approach can be used on other image data sets that contain a large number of similar objects, such those in biology, satellite imagery, astronomy and material sciences.

Finally, it is straightforward to extend the idea of C2G-Net to 3-dimensional data. For example, 2-dimensional images of consecutive sections of FFPE tissue (with slice thickness of $\approx$\SI{4}{\micro\metre} \cite{Robertson2008-pv}) can be compressed with Cell2Grid by using a target grid spacing of the same width. By virtual stacking of these images, a 3-dimensional data model containing each object as a cube could be created. We hypothesize that training a 3-dimensional CNN on this data could further improve model performance.

\section{Conclusion}
\label{conclusion}
In this paper, we investigated if the morphological properties of images containing a large number of similar objects like biological cells can be exploited for image classification.
For this purpose, we introduced C2G-Net consisting of Cell2Grid, an image compression algorithm, and DeepLNiNo, a CNN architecture aimed at facilitating model interpretability.

As a case study, we investigated predicting relapse risk in colon cancer using multiplex immunohistochemistry images.
We found that Cell2Grid alone improved prediction accuracy of a CNN compared to conventional image compression.
Moreover, when comparing the entire C2G-Net pipeline to models trained on high resolution raw images, C2G-Net showed comparable prediction accuracy, reduced training time by more than 85\%, and provided a model that is easier to interpret.



\bibliography{c2g-net}
\bibliographystyle{icml2020}

\appendix

\newpage
\section{PriorityShift: resolving assignment conflicts}
\label{AppendixA}

During step 2 of Cell2Grid image compression, i.e. binning of objects to the target grid, several objects might be assigned to the same grid node $\vec{X}_g$. These conflicts become more frequent with increasing grid spacing $d$. A finer grid spacing might prevent such conflicts but also reduces the compression ratio of the algorithm. In any case, assignment conflicts need to be resolved to obtain a one-to-one mapping of objects to pixels. This can be achieved either by deleting conflicting objects or moving them to adjacent free grid nodes. Algorithm \ref{alg:priorityshift} introduces \textit{PriorityShift}, a simple and fast method that resolves assignment conflicts while minimizing the amount of objects that need to be deleted.

To make this algorithm computationally efficient, we reduced the number of times that euclidean distances need to be calculated. For $n$ conflicting objects, we only calculate $n$ euclidean distances from each object to the conflicting grid node $G$. When determining which adjacent grid node is closest to an object, we exploit the geometry of the square grid by calculating in which of the eight quadrant halves around $G$ the object is located. This location directly determines the priority of adjacent grid nodes. The conflicting object is then shifted to the free grid node with highest priority.

\begin{algorithm}[hbt]
   \caption{PriorityShift}
   \label{alg:priorityshift}
\begin{algorithmic}
   \STATE {\bfseries Input:} grid with assigned objects containing assignment conflicts
   \FOR{every grid node $G$ with conflicting objects}
   \STATE get original locations of objects conflicting at $G$
   \STATE sort conflicting objects by proximity to $G$
   \STATE get list $L$ of free grid nodes directly adjacent to $G$
   \STATE assign the object closest to $G$ to grid node $G$
   \FOR{remaining conflicting objects at $G$}
   \IF{L contains at least one element}
   \STATE determine priority of grid nodes in $L$
   \STATE assign objects to node in $L$ with highest priority
   \STATE update $L$
   \ELSE
   \STATE delete object from the data
   \ENDIF
   \ENDFOR
   \ENDFOR
   \STATE {\bfseries Output:} grid with assigned objects without conflicts
\end{algorithmic}
\end{algorithm}


\newpage
\section{Data augmentation for Cell2Grid images}
\label{AppendixB}

To prevent CNNs from overfitting \cite{Wong2016-bk,Taylor2017-qe}, a multitude of data augmentation methods for conventional images are available, including translations, rotations, zooming, shearing and others \cite{Chollet2015-nn,Bloice2019-km}. However, Cell2Grid images are different from conventional images, as outlined in the main text. Therefore, traditional image augmentation methods are not applicable or need to be modified. On the other hand, new augmentation methods specific to Cell2Grid images can be used.




We used the following data augmentation methods for Cell2Grid images of size \SI{135x101}{\px}. Each method was applied with probability $p$ to an image during training:

\begin{itemize}
\item \textit{Translations} ($p=100\%$). We translated images horizontally (max. \SI{30}{\px}) and vertically (max. \SI{20}{\px}).
\item \textit{Reflections} ($p=100\%$). We performed horizontal and/or vertical reflections.
\item \textit{Discrete rotations} ($p=100\%$). We rotated images around \SI{0}{\degree}, \SI{90}{\degree}, \SI{180}{\degree} or \SI{270}{\degree}.
\item \textit{Blackouts} ($p=80\%$). We deleted all objects in an area of \SI{25x25}{\px} at a random location.
\item \textit{Local pixel shuffle} ($p=100\%$). For 50 random locations, we shuffled the objects in an area of \SI{3x3}{\px}.
\item \textit{Channel brightness change} ($p=10\%$). We changed the brightness of each color channel individually by multiplication of all values with a random factor in the range $[0.9, 1.1]$.
\item \textit{Global brightness change} ($p=100\%$). We changed the brightness of all channels together with a random factor in the range $[0.8, 1.2]$.
\item \textit{Deleting pixels} ($p=100\%$). We randomly chose 100 pixels and set their values to 0.
\end{itemize}

Where applicable, we used the "reflect" mode of Keras to fill empty pixels, otherwise we filled occurring empty pixels with $0$.

\end{document}